\documentclass[
  superscriptaddress,
  amsmath,
  amssymb,
  aps,
  reprint,
  prx,
  twocolumn,
  nofootinbib,
  longbibliography,
  floatfix
]{revtex4-2}

\usepackage[utf8]{inputenc}
\usepackage[T1]{fontenc}
\usepackage{amsfonts,amssymb,amsmath,mathtools,amsthm,bm}
\usepackage{graphicx}
\usepackage{booktabs}
\usepackage{nicefrac}
\usepackage{microtype}
\usepackage{xcolor}
\usepackage{xurl}
\usepackage[colorlinks=true,allcolors=black]{hyperref}
\hypersetup{
  pdftitle={Meta-memorization and memorization scaling laws in transformers},
  pdfauthor={Alex Nguyen, Kenneth A. Norman, Gautam Reddy}
}

\theoremstyle{plain}

\theoremstyle{definition}

\theoremstyle{remark}

\begin{document}
\raggedbottom

\title{Meta-memorization and memorization scaling laws in transformers}

\author{Alex Nguyen}
\affiliation{Princeton Neuroscience Institute, Princeton University, Princeton, NJ, USA}
\author{Kenneth A. Norman}
\affiliation{Princeton Neuroscience Institute, Department of Psychology, Princeton University, Princeton, NJ, USA}
\author{Gautam Reddy}
\affiliation{Department of Physics, Princeton University, Princeton, NJ, USA}

\begin{abstract}
Larger transformers can memorize more data, with a capacity (in bits) that scales roughly linearly with the number of parameters. Counterintuitively, larger models also memorize \emph{faster}, requiring fewer repetitions of a sequence during training before it is memorized. We reproduce both memorization scaling laws in transformers trained on random binary sequences, showing that rapid memorization can arise even when data contain no semantic or compositional structure. Rapid memorization is due to a \emph{meta-memorization} effect, where models trained to memorize a large set of random sequences learn to memorize novel sequences more quickly. The rate $r(t)$ at which models memorize grows with training time $t$ as a power-law $r(t) \sim t^{\beta}$ with a fixed exponent $\beta$ but an increasing upper cutoff across model scales. Meta-memorization enables even one-layer transformers to few-shot memorize a 100-bit sequence. This effect is partly explained by the formation of structured attention heads that convey position-invariant sequence information to MLPs where memorized sequences are stored. Meta-memorization is recapitulated in a simplified model of a dense associative memory where retrieval is constrained by an attentional bottleneck. Our results suggest transformers, through pre-training, acquire reusable mechanisms that enable highly efficient storage and retrieval of memories. 
\end{abstract}

\maketitle

\section{Introduction}
\label{submission}

Significant attention has been paid to the ability of transformers to generalize with limited data, but transformers are also remarkable memorizers ~\citep{vaswani2017attention,lampinen2024broader}. Large language models (LLMs) can reproduce training passages verbatim when prompted with sufficient context and this can be exploited to extract training data ~\citep{carlini2019secret, carlini2022quantifying, nasr2023scalable}. For example, production language models verbatim recall much of \emph{Harry Potter and the Sorcerer's Stone} when prompted with the first few tokens of the first chapter~\citep{ahmed2026extracting, cooper2025extracting}. 

One influential approach to quantify memorization is based on extractability. Given a prefix of a training example, one can ask whether the model reproduces its exact continuation, and measure how this probability changes with context length, dataset duplication, and scale~\citep{biderman2023emergent,carlini2022quantifying, hayes2025measuring,kandpal2022deduplicating}. A complementary approach measures memorization dynamically during training, asking how many presentations are necessary before examples become exactly reproducible~\citep{huang2024demystifying,tirumala2022memorization}. 

Several robust features have emerged from these measurements. Multiple analyses of pretrained transformers suggest that MLP layers store associations, consistent with the idea that MLP blocks function as key-value memories~\citep{dong2025attention, geva2021transformer, geva2022transformer, geva2023dissecting, meng2022locating, nichani2024understanding}. Memorization increases with an LLM's parameter count and with the number of times a sequence is duplicated in the training corpus, often in simple log-linear relationships~\citep{carlini2022quantifying,chen2026comparative}. More broadly, recent work shows that transformers' storage capacity (measured in bits, after controlling for generalization) scales linearly, with a few bits stored per parameter~\citep{allen2024physics, morris2025much, nichani2024understanding}. Training proceeds through regimes in which the model first fills this capacity before shifting toward more reusable structure \citep{zucchet2025language,zucchet2025emergence}. 

Counterintuitively, memorization does not just scale in capacity but also in rate; that is, larger models require fewer presentations of a sentence before it is memorized~\citep{tirumala2022memorization, carlini2022quantifying}. Having more parameters can explain greater capacity, but this does not explain why incremental gradient updates should become systematically more effective at creating a memory. Resolving this puzzle in language models is difficult as memorization rate is inherently tied to the model’s improving ability to exploit the structure of language~\citep{panwarbetter}. 
\\

\noindent \textit{Contributions}. In this work, we study how transformers memorize random binary sequences that have no internal statistical structure. We reproduce several empirical results in the literature, in particular that both transformers' memorization capacity and rate increase with size \citep{carlini2022quantifying,tirumala2022memorization}, derive a scaling law in the memorization rate as a function of model size, and show that large models can achieve one-shot memorization. Training models to memorize random sequences reduces the number of presentations required to memorize any arbitrary new sequence, an effect we call \textit{meta-memorization} (in connection with meta-learning). We show that transformers meta-memorize through a functional division of labor, where attention heads convey position-invariant, linearly separable information about sequence identity and the MLPs store dataset-specific memories. Finally, we instantiate our theory of the distinct functional roles of attention and MLPs in a conceptual associative memory model, where attention heads convey information through a constrained bottleneck to the associative memory. The code for reproducing experiments and analysis can be found in the following  \href{https://github.com/anhhuyalex/icl-memorization}{\textcolor{blue}{repository}}.

\section{Setup}
\begin{figure}[!tbp]
     \centering
     \includegraphics[width=\linewidth]{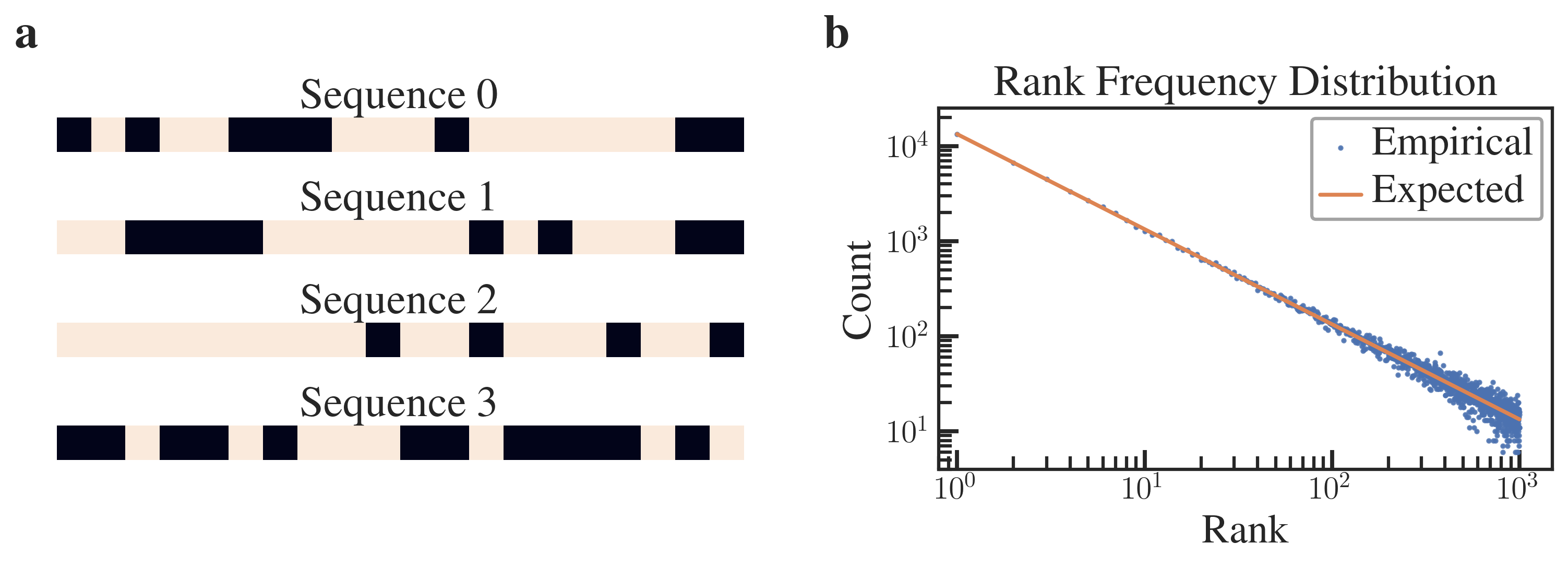}
     \caption{Experimental setup. We train autoregressive transformers on $K=10^5$ random binary sequences of length $N=100$. (a) Four example sequences. (b) Sequences are assigned ranks and sampled from a Zipfian distribution, producing repetition counts spanning orders of magnitude. For illustrative purposes we only plot up to rank $10^3$. Model size is varied through the number of heads $H$ and model dimension $D$.
     }
     \label{fig:setup} 
 \end{figure}

Transformers of different sizes are trained autoregressively on a dataset of randomly generated binary sequences (Figure~\ref{fig:setup}a). The dataset is constructed with a Zipfian rank-frequency distribution (Figure~\ref{fig:setup}b). Specifically, we sample $K=10^5$ random binary sequences of length $N=100$ and assign each sequence a rank $k$. Each bit in each sequence is sampled i.i.d.\ from a Bernoulli distribution ($p = 0.5$) (Figure~\ref{fig:setup}a). During training, a sequence of rank $k$ is sampled from a Zipfian distribution $p(k) \propto k^{-1}$ (Figure~\ref{fig:setup}b). Given the first $n$ bits, the model predicts the $(n+1)$-th bit under a binary cross-entropy objective. 

This design serves three purposes. First, since the sequences are unstructured, it inherently controls for any unintended memorization due to generalization. Second, it mirrors a salient, domain-general feature of natural data that is known to influence emergent transformer behaviors in other contexts~\citep{chan2022data}. Third, it allows us to ask whether the rate at which a novel (or a less frequent) sequence is memorized depends on the model’s prior experience memorizing other more common sequences. 

We train decoder-only transformers with learnable positional embeddings and causal masking. We consider two classes of models: (1) GPT-2 style transformers with depth $12$~\citep{radford2019language,karpathy2022nanogpt}, and (2) one-layer models with one layer of attention followed by a three-layer MLP. In both cases, model size is varied by varying the number of heads ($H$) and the per-head width ($F$), which in turn fixes the model dimension ($D = FH$). The 12-layer models are trained for $2 \times 10^5$ iterations, whereas the 1-layer models are trained for $8 \times 10^4$ iterations since they converge more quickly. All models are trained with an Adam optimizer (learning rate $10^{-3}$ and batch size $256$)~\citep{kingma2014adam}.

For each sequence, we compute the average next-token prediction accuracy by feeding the model the sequence prefix and comparing the predicted next token to the ground truth. Accuracy is averaged across all positions. A sequence is considered ``memorized'' when the model achieves 90\% average per-token accuracy. For a given model and a sequence of rank $k$, we measured the training iteration $\tau_k$ at which the model first reached this threshold, and the total number of sequence presentations $n_k$ the model had received up to iteration $\tau_k$. The use of both measures allows us to disentangle the effects of training iterations from sequence frequency.

\section{Results}\label{sec:results}
\subsection{Memorization scaling laws and meta-memorization}

\begin{figure*}[!t]
     \centering
     \includegraphics[width=0.95\textwidth]{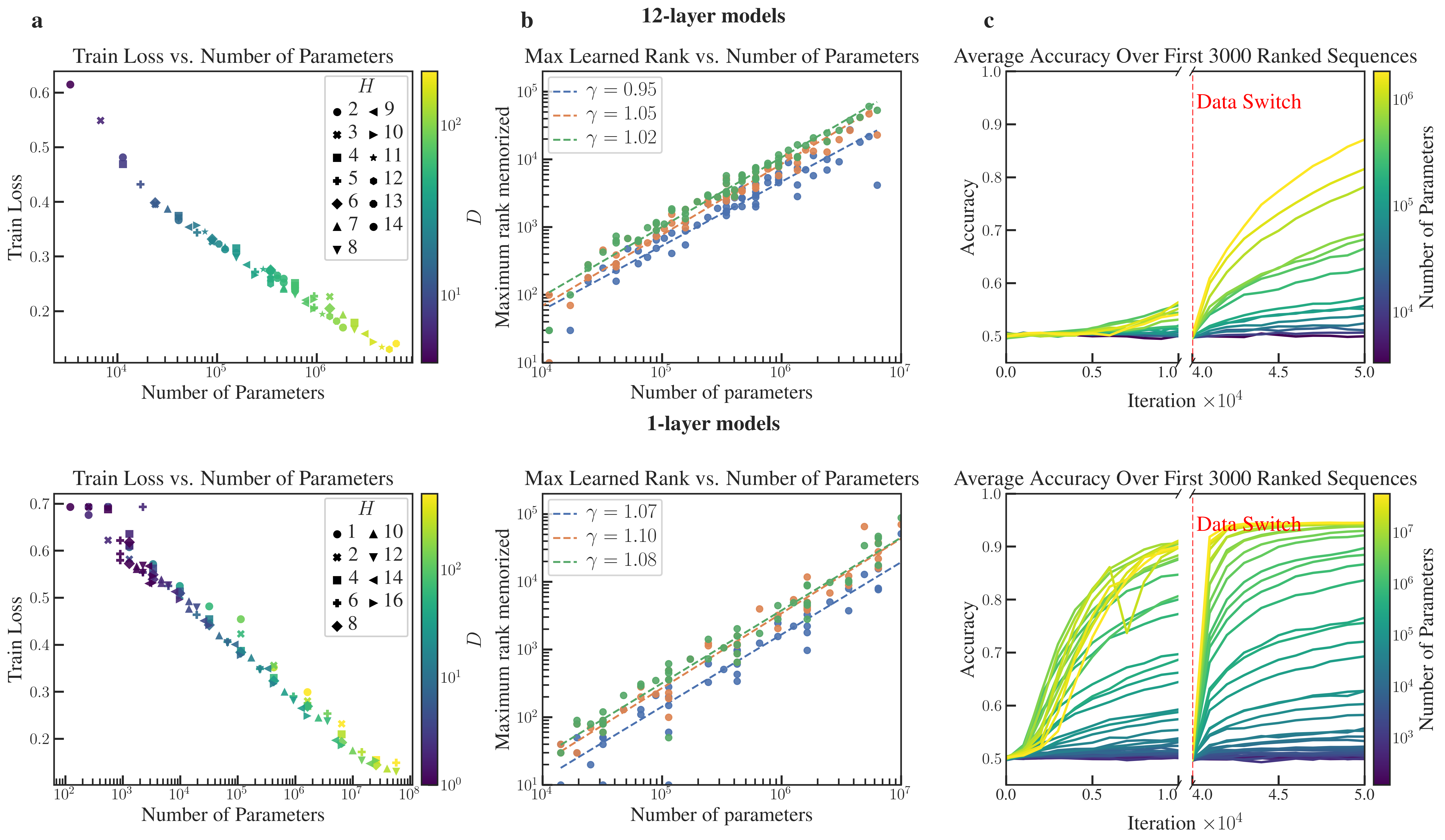}
     \caption{Memorization scaling laws and meta-memorization for 12-layer and 1-layer architectures. (a) For both architectures, final train loss decreases primarily with total parameter count of models even though models vary in both number of heads $H$ and model dimension $D$. The 12-layer and 1-layer models are trained for $200,000$ and $80,000$ iterations, respectively, with a fixed learning rate. Note that the largest models failed to converge with this learning rate and are not shown in the plot. (b) The maximum rank of memorized sequences ($\propto$ the memorization capacity $C$ in bits) shown at $60,000$, $120,000$ and $190,000$ iterations for the 12-layer models and $20,000$, $50,000$ and $79,000$ iterations for the 1-layer model. The exponent of a power law fit $\gamma$ is $\approx 1$ for both architectures and during training, thus showing that memorization capacity $C$ scales linearly with parameter count $P$. (c) Larger models memorize sequences faster. Moreover, models memorize data from a second dataset faster after a dataset switch (red dashed line). Due to the rapid learning in 1-layer models, here we show results only for iterations $[0,10^4]$ and $[4\times10^4,5\times 10^4]$ to better illustrate the effect.}
     \label{fig:perf} 
 \end{figure*}

For both 12-layer and 1-layer models, the average loss and accuracy across models with different numbers of heads $H$ and model dimensions $D$ collapse onto a single curve that depends on architecture solely through the total parameter count $P$ (Figure~\ref{fig:perf}a, Figure~\ref{fig:acc_scaling}). As expected, larger models generally show lower train loss and higher recall of the most frequent sequences. The dependence of loss and accuracy on parameter count was also recapitulated in separate experiments with uniformly-drawn (rather than Zipfian) sequences (Figure~\ref{fig:uniform_loss_accuracy}). 
\\

\noindent \textit{Memorization capacity scaling law.} We define memorization capacity ($C$) as the total number of bits memorized by the model after convergence. Because models, to a good approximation, memorize sequences sequentially by rank, the maximum memorized rank $M$ (which is easily measured) tracks the number of memorized sequences. In all capacity estimates below we therefore use $M$ as a proxy for the number of memorized sequences. 
Since each sequence is 100 bits, we have $C \approx 100M$ bits. We measure $M$ vs $P$ at three different points during training as the loss continues to decline gradually at the end of training (Figure~\ref{fig:loss_dyn}).   For both model classes, $M$ scales with parameter count as $M \sim P^{\gamma}$ (Figure~\ref{fig:perf}b). We find that $\gamma \approx 1$  with a pre-factor $\approx$ $5\times 10^{-3}$ and $10^{-2}$ for 1-layer and 12-layer models, respectively. In other words, memorization capacity scales linearly with parameter count, consistent with previous measurements. Our measurement leads to an estimate of $C/P \approx 0.5-1$ bits per parameter.
\\

\noindent \textit{Meta-memorization.} By measuring the training iteration $\tau_k$ at which a sequence with rank $k$ is memorized, we find that larger models memorize sequences (of a given rank) earlier in training (Figure~\ref{fig:saturation_raw}a).  That is, larger models appear to memorize sequences faster.  Re-plotting this data using number of presentations $n_k$ after which a sequence of rank $k$ is memorized over all models shows that $n_k$ generally decreases with rank  (Figure~\ref{fig:saturation_raw}b). Under Zipfian sampling, high-rank sequences are seen less frequently and therefore tend to cross the threshold later during training. If they nevertheless require fewer presentations to be memorized, then the effective amount of memory written per presentation must have increased over the course of training. We term this effect \emph{meta-memorization}, where memorizing a set of sequences allows a model to memorize new sequences more efficiently. We suggest the term meta-memorization to evoke the connection to meta-learning, where training on a task distribution improves model performance on other similar tasks. Our finding replicates \cite{tirumala2022memorization}'s finding that larger models memorize linguistic data faster, but importantly, we show that this memorization rate scaling law holds even when the training data lack any statistical structure.
\\

\noindent \textit{Data switch experiments.} To test this directly, we performed an experiment where the training set of sequences was switched after $4\times 10^4$ iterations to a fresh set of random sequences sampled from the Zipfian distribution. Accuracy immediately returned to chance after the switch, but recovered substantially faster than it had at the start of training (Figure~\ref{fig:perf}c). The speedup was visible for sequences across all ranks (see Figure \ref{fig:12layer_data_switch_3_panels} for 12-layer models and Figure \ref{fig:OL_data_switch_3_panels} for 1-layer models). 
\\

\noindent \textit{Memorization rate scaling law.} Next, we estimate how memorization rate evolves during training. Suppose a sequence of rank $k$ is sampled with probability $p_k$ and first crosses the memorization threshold at iteration $\tau_k$. 
Let $r(t)$ denote the effective memorization rate at training time $t$ in units of bits per iteration. The extent to which sequence $k$ is memorized (in bits) at $t$ is $m_k(t) = \sum_{s \le t} r(s) \delta_{k,\iota(s)}$, where $\iota(s)$ is the rank of the sequence that was presented at iteration $s$. A sequence of $N$ bits is memorized when $m_k(\tau_k) = N$. Let $R(t)=\sum_{s \leq t} r(s)$. If $n_k \gg 1$, we have
\begin{align}
    m_k(\tau_k) \approx p_k R(\tau_k) \approx N. \label{eq:mk}
\end{align}
Suppose the sequences are drawn according to a rank-frequency distribution $p_k \sim k^{-\rho}$, and that $n_k \sim k^{-\alpha}$ (or equivalently $\tau_k \approx n_k/p_k \sim k^{-\alpha + \rho}$) for some $\alpha$. Empirically, we find $\alpha > 0$ (Figure~\ref{fig:saturation_raw}b). Under the ansatz $r(t) \sim t^{\beta}$ for some $\beta$, we obtain $R(t) = \sum_{s \leq t} r(s) \sim t^{\beta+1}$. Combining \eqref{eq:mk} and these scaling relationships, we get
\begin{align}
    \beta = \frac{\alpha}{\rho - \alpha}. \label{eq:beta}
\end{align}
\begin{figure*}[!t]
     \centering
     \includegraphics[width=0.9\textwidth]{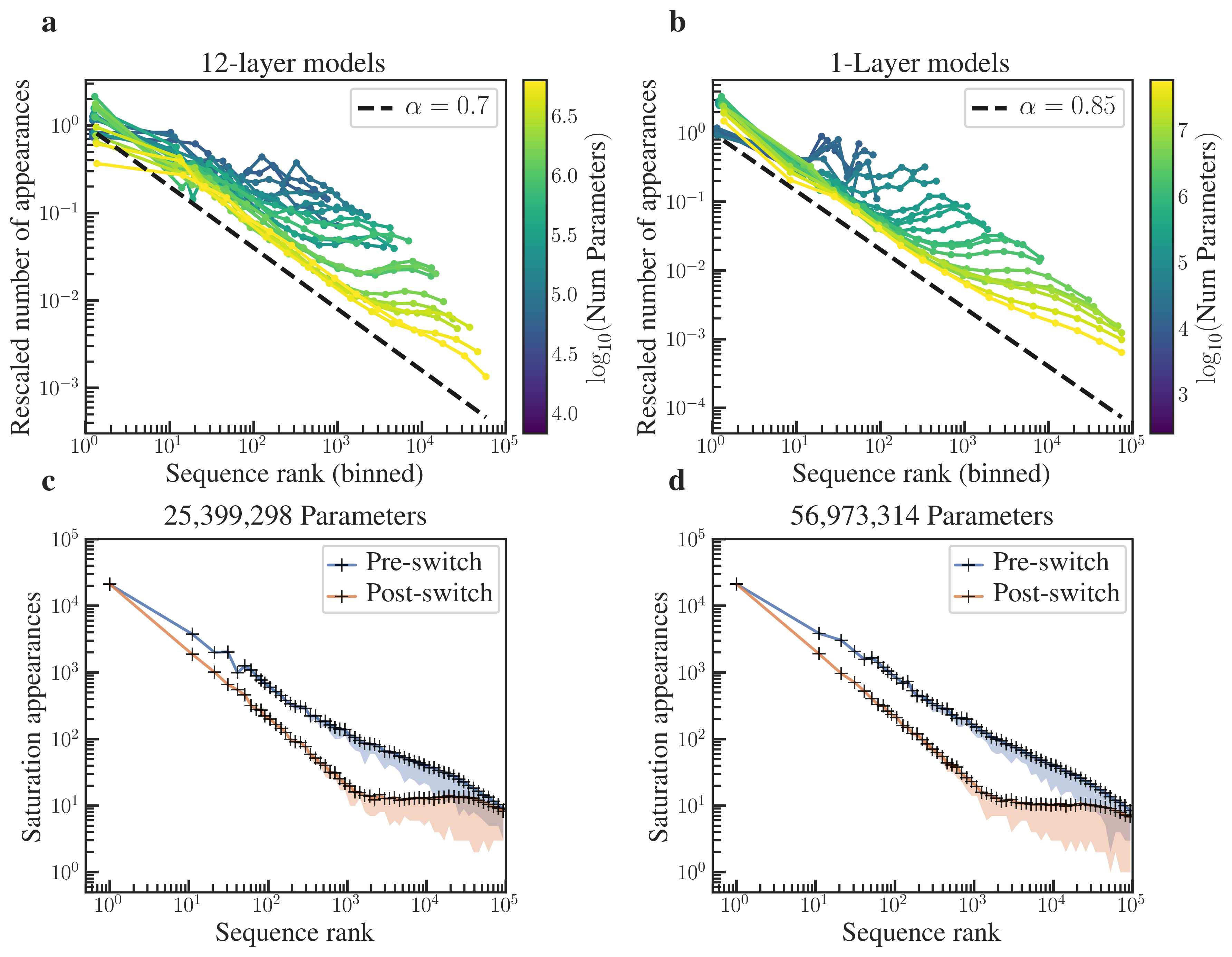}
     \caption{Meta-memorization scaling laws. Across models, we find that fewer presentations are required to memorize a rarer sequence (of higher rank) (Figure~\ref{fig:saturation_raw}b). Since higher-rank sequences tend to appear later on during training, this suggests that models become more efficient at memorizing over the course of training.  In panels a) and b), the number of presentations ($n_k$) vs rank ($k$) data is first fit to a power-law $n_k = Ak^{-\alpha}$ for each model. We plot the rescaled number of presentations (i.e., after dividing by $A$ for each model) vs rank for all models. Across model sizes, 12-layer and 1-layer models display power laws with exponents  $\alpha \approx 0.7$ and $\alpha \approx 0.85$, respectively. The exponent is common to all model sizes within a model class, but the scaling extends to higher ranks for larger models. In panels c) and d) we show appearances at saturation vs. sequence rank in data-switching experiments for the largest 1-layer models. Appearances at saturation is the number of presentations $n_k$ at which a sequence $k$ is memorized (i.e. learned to 90\% accuracy). To avoid visual clutter, we bin the x-axis logarithmically and show the geometric mean of sequence ranks and appearances at saturation within each bin. Error bands show the minimum number of appearances among ranks that fall in each logarithmic bin. We show these pre-switch and post-switch data and observe that after the switch, sequences of equal rank require fewer presentations to be learned. (d) Furthermore, the largest model at $5.7\times 10^7$ parameters was able to learn some sequences in \textit{one-shot} after the switch (the error band reaches 1 appearance after the switch).}
     \label{fig:scaling_law} 
 \end{figure*}
$\beta$ is estimated by fitting $n_k$ vs $k$ to a power law (recall $\rho = 1$ for the Zipfian rank-frequency distribution). Remarkably, we find that a fixed $\alpha \approx 0.7$ and $\alpha \approx 0.85$ fits the data from the 12-layer and 1-layer models, respectively, across all model scales (Figure~\ref{fig:scaling_law} panels a and b). Model sizes are primarily differentiated by the rank at which $n_k$ stops decreasing (Figure~\ref{fig:scaling_law} panels a and b). In other words, $r(t) \sim t^{\beta}$ for all models (with an almost unvarying $\beta$), but the range of $k$ for which the scaling law applies is larger for larger models.  Using Eq. \eqref{eq:beta}, we estimate $\beta \approx 2-6$ across the 12-layer and 1-layer model classes (the estimate has a wide range as $\alpha$ is close to 1). Thus, the memorization rate accelerates during training. In the extreme case, the largest 1-layer models are able to memorize a novel sequence after a single presentation of the sequence (Figure \ref{fig:scaling_law} panel d).

\begin{figure*}[!t]
\centering
\includegraphics[width=0.9\textwidth]{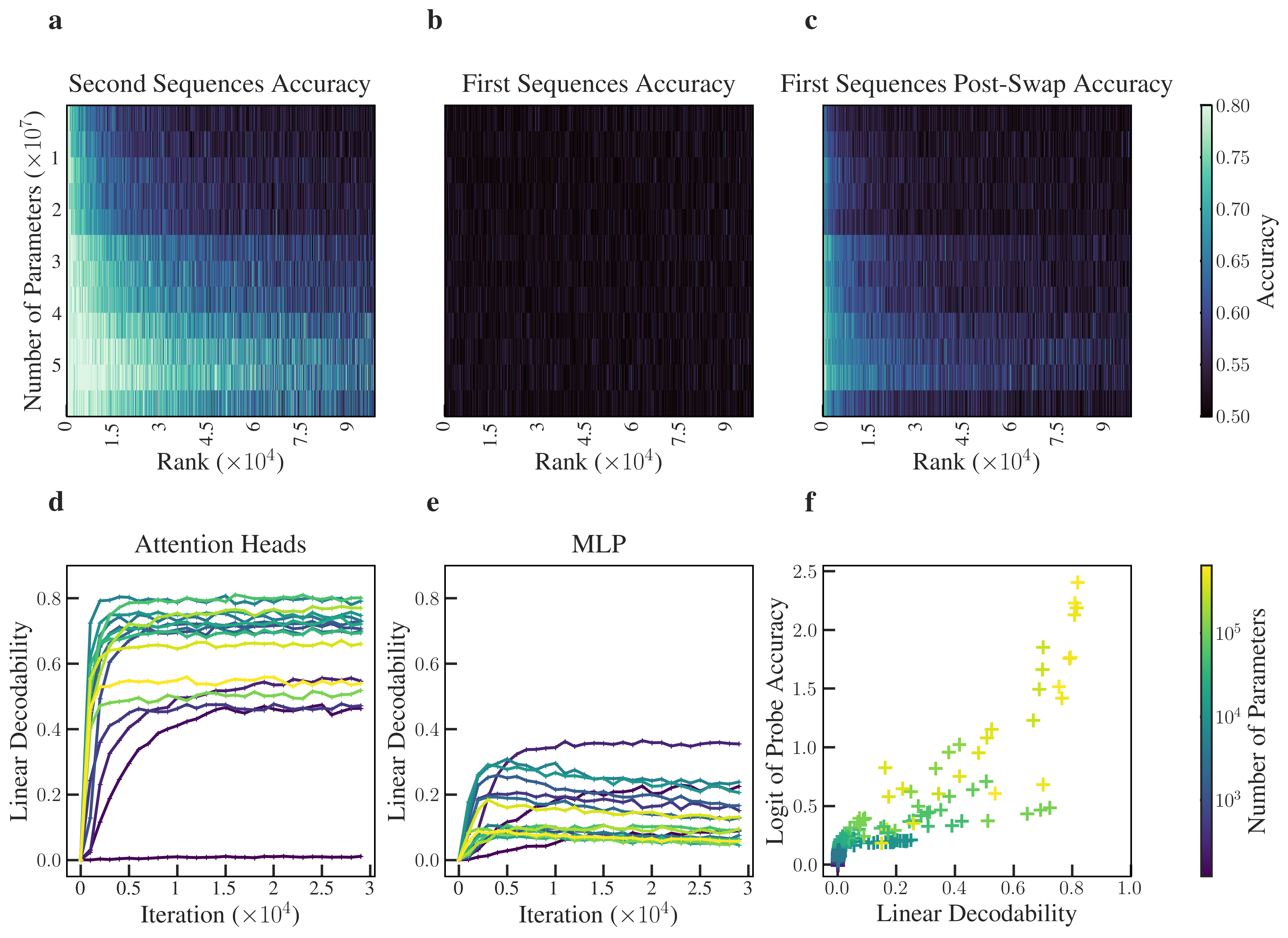}
\caption{Division of labor among feedforward blocks and attention heads. \textbf{Top}: We train models where half-way through training, we resample all training sequences. (a) Accuracy of models after training on second sequences as a function of model size and sequence ranks. (b)  Accuracy of models on first sequences as a function of model size and sequence ranks. (c) Accuracy of component-swapped models on first sequences, using MLPs at the end of the first period. \textbf{Bottom}: Linear decodability of 1-layer representations. (d) Linear decodability of 1-layer attention representations of novel sequences unseen during training as a function of training iteration. (e) Linear decodability of MLP representations of novel sequences unseen during training as a function of training iteration. (f) Linear decodability of attention representations predicts accuracy of probe experiments.}
\label{fig:mlps_memorize}
\end{figure*}

\subsection{The division of labor between attention heads and feedforward blocks}

The one-layer architecture allows for a deeper analysis of the differential roles of attention and feed-forward components in memorization, while still capturing the essential phenomenology. We perform a series of experiments motivated by previous work on factual recall that suggests facts are localized in feedforward blocks (MLPs) \cite{geva2021transformer, dong2025attention, meng2022locating}. 
\\

\noindent \textit{Component swapping.} First, we conducted a component-swapping experiment, building on the data switch experiments discussed previously. Recall that the data switch experiments consist of models trained for $4 \times 10^4$ iterations, and subsequently trained for another $4 \times 10^4$ iterations on a newly generated set of sequences. We saved model weights at two timepoints: immediately before the switch ($\theta_{\text{pre}}$) 
 and at the end of training ($\theta_{\text{post}}$). If the MLP stores sequence-specific associations, then patching in the weights of an MLP saved before a dataset switch should rescue recall accuracy on the pre-switch set of sequences even when paired with post-switch attention weights.

As expected, the final model ($\theta_{\text{post}}$) achieved high accuracy on the new set of sequences (Figure~\ref{fig:mlps_memorize}a) but has catastrophically forgotten the original set (Figure~\ref{fig:mlps_memorize}b). We then constructed a hybrid model combining the attention weights from $\theta_{\text{post}}$ with the MLP weights from $\theta_{\text{pre}}$, denoted $\theta_{\text{hybrid}} =(\text{Attn}_{\text{post}}, \text{MLP}_{\text{pre}})$. We evaluated this hybrid on the original set without any additional training. If sequence-specific information were stored primarily in attention weights, then substituting the post-switch attention should eliminate recall of the original sequences. Conversely, if sequence-specific associations reside in the MLP, the pre-switch MLP weights should enable above-chance performance on the original set, even when paired with attention weights associated with a model that has catastrophically forgotten the original set. Figure~\ref{fig:mlps_memorize}c shows that the hybrid model $\theta_{\text{hybrid}}$ indeed achieves above-chance accuracy on the original set. 
\begin{figure*}[!t]
    \centering
    \includegraphics[width=0.95\textwidth]{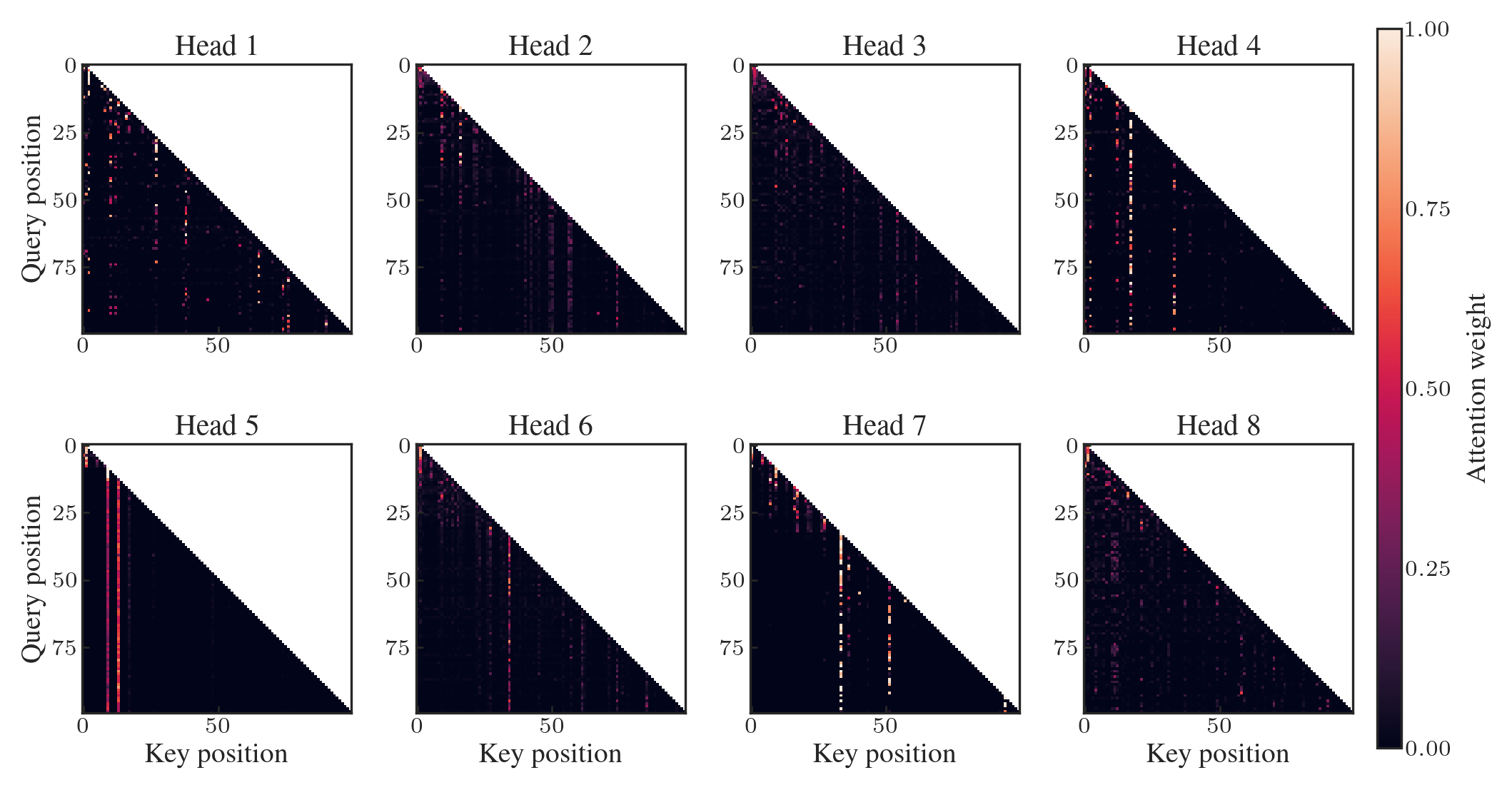}
    \caption{Learned attention maps in a trained one-layer transformer. Each panel shows one of 16 heads with $F=64$ features per head, evaluated on the most frequent sequence. Rows are query positions and columns are causal key positions. Vertical bands indicate that heads specialize to fixed positions or position subsets, providing reusable context features to the MLP at each position.  See Figure \ref{fig:attention_weights_sequence_10} and Figure \ref{fig:attention_weights_head_5} for attention maps for a different sequence and for a single head.}
    \label{fig:attention_maps}
\end{figure*}
\\

\noindent \textit{Linear decodability.} Next, we hypothesize that attention outputs preserve information about sequence content by developing linearly separable representations of the sequence, and that the MLP transforms this information into predictions about the next bit. To quantify how explicitly sequence identity is represented in different layers, every 1000 iterations, we presented 1000 novel sequences to the model and extracted activation vectors from both attention outputs and MLP outputs at 25 evenly spaced positions across sequences of length $N=100$ (positions 0, 4, 8, \ldots, 96). This yielded a classification dataset where the task is to predict sequence identity from activations at a given position. We trained separate logistic classifiers for attention and MLP representations at regular intervals throughout model training. 

We observed that attention representations rapidly developed highly linearly separable representations of sequence identity: linear classifier accuracy exceeded 80\% early in training and continued to improve (Figure \ref{fig:mlps_memorize}d). While MLP classifier accuracy initially increased, it plateaued well below 50\% accuracy and \textit{decreased} over the course of training (Figure \ref{fig:mlps_memorize}e). We hypothesized that this decline is because the MLP must focus on predicting the next bit rather than maintaining sequence identity information. 
\\

\noindent \textit{Probe fork experiments.} To further show that linearly separable attention representations benefit memorization, we train 1-layer models where every 50 iterations during training, we initialize a new model with the same weights as the current model, freeze the attention weights and train \emph{only} the MLP weights. We then ``fine-tune'' this forked model exclusively on a held-out set of probe sequences (never seen during training) which are presented uniformly at random. We fine-tune for up to 1,000 iterations, stopping early if the model obtains 90\% accuracy.  

For each 1-layer model, we plot the test accuracy of the linear classifier at various times during training against the logit-transformed fine-tuning accuracy and find that more linearly separable attention representations predict better fine-tuning accuracy (Figure \ref{fig:mlps_memorize}f). Figure \ref{fig:mlps_memorize}f also shows that larger models attain higher accuracy during fine-tuning. The fine-tuning accuracy of 1-layer models also increases throughout training, with the effect most apparent for larger models (Figure \ref{fig:finetuning_accuracy_vs_time}).
\\
\begin{figure*}[!t]
    \centering
    \includegraphics[width=0.95\textwidth]{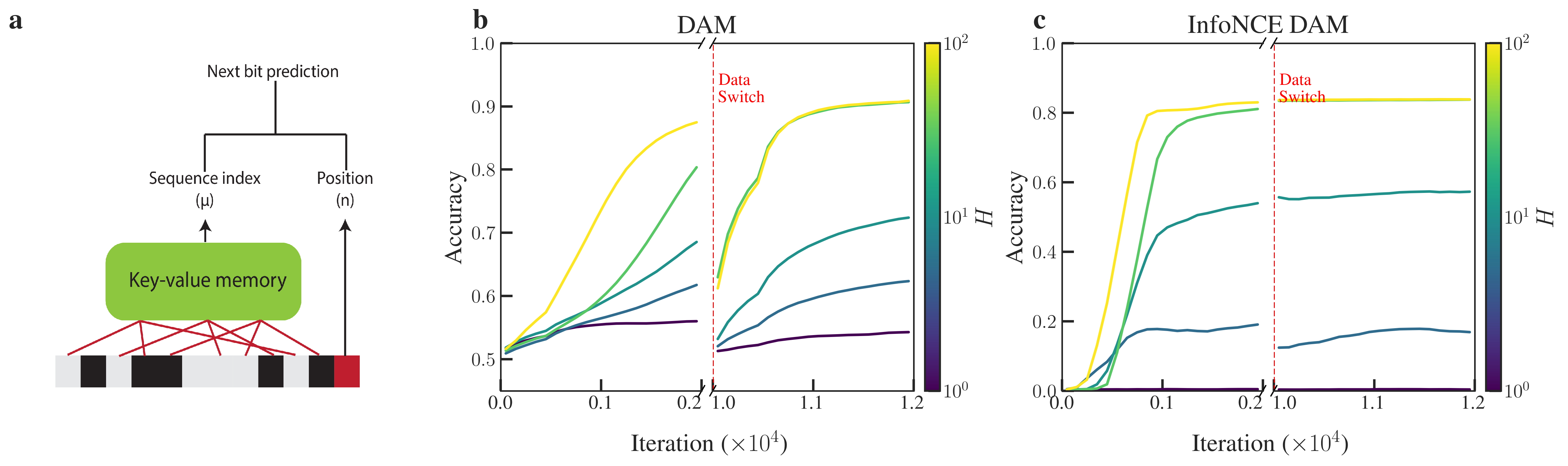}
    \caption{A phenomenology-driven theory of meta-memorization. (a) In this model, the attention heads present position-invariant information about the sequence to a key-value memory through a finite attentional bottleneck. The key-value memory uses this sequence information to first retrieve sequence index ($\mu$) and a downstream decoder uses sequence index $\mu$ and the current position in the sequence $n$ to predict the $n+1$th bit.  (b) The DAM model recapitulates the finding that switching to new random sequences reveals meta-memorization. Here, $A$ and $B$ are trained continuously under the next-bit objective. Accuracy drops when the sequence set is replaced but recovers faster than during initial training. We show results only for iterations $[0,2\times 10^3]$ and $[10^4,1.2\times 10^4]$ to better illustrate the effect. (c) Here, $A$ and $B$ are trained using an InfoNCE-style retrieval objective. The model shows post-switch accuracy comparable to pre-switch accuracy even when $A$ and $B$ are frozen after the switch. In both cases, transfer strengthens with the number of heads $H$. Note that the noisy accuracy curves (due to finite-batch effects) are smoothed by computing a running average. This explains why the accuracy does not appear to return to random chance in panel b after the data switch, since the first point depicted is a running average of the first 100 steps' accuracies after the switch. Here, we use $N=50,M=3000,K=500$, batch size = 50.}
    \label{fig:dam}
\end{figure*}

\noindent \textit{Homogenized attention.} We verified that models with attention weights set to zero throughout training (such that attention heads provide no position-specific contextual information) failed to learn the training sequences and, correspondingly, forks showed little to no improvement in fine-tuning accuracy regardless of training duration (Figure \ref{fig:zero_attn_weights}). 
\\

\noindent \textit{Attention maps.} Finally, we asked if attention heads acquire particular structure that allows them to better represent sequence identity. To do this, we visualize attention maps for a trained one-layer model with 16 heads (64 features per head) in Figure \ref{fig:attention_maps}. Here we compute the attention paid at each query position to the prior positions for Sequence 1 (the highest rank and thus most commonly presented sequence). Sample attention maps for Sequence 10 are presented in Figure \ref{fig:attention_weights_sequence_10} and the fifth head's attention maps in Figure \ref{fig:attention_weights_head_5}. Across heads and sequences, we observe prominent vertical bars in the attention maps, indicating that each attention map specializes to read from a particular set of positions. The vertical stripe patterns can be intuitively understood: each head conveys the same information to the MLP regardless of the length of the current subsequence. This allows the MLP to learn a position-independent decoder of sequence identity.

\subsection{A theory of meta-memorization}
Put together, these results support a functional division between attention and MLP layers in transformers such that attention heads convey sequence information to MLPs where memories are stored. This picture is consistent with prior work, notably \cite{zucchet2025language}, who show that attention circuits develop before MLPs begin to store factual information. Note that their task involves memorization of random synthetic facts expressed in natural-language which retains generalizable linguistic structure, making it challenging to identify what is learned by the attention heads. Our results show that attention heads learn to construct position-invariant, linearly separable representations of sequence identity and the MLP transforms these representations into dataset-specific features optimized for next-token prediction. 
\\

\noindent \textit{Conceptual picture.} Based on our results and prior work, we propose a theory of meta-memorization where the attention layers and feedforward blocks in a transformer specialize towards performing specific computational roles over the course of training (Figure~\ref{fig:dam}a). Specifically, we propose the attention heads convey position-invariant information about sequence identity to a key-value memory, which stores sequences encountered during training. During evaluation, the key-value memory uses sequence information conveyed by attention heads to first retrieve sequence identity (or index, denoted $\mu$). A downstream decoder then uses sequence index $\mu$ and the current position in the sequence $n$ to predict the $n+1$th bit. The amount of sequence information that can be conveyed to the key-value memory is limited, since the number of attention heads $H$ and the number of features conveyed by each head $F$ are finite. The attentional bottleneck is necessary to encourage attention heads to learn an efficient coding scheme. Importantly, this bottleneck explains why larger models learn faster, since the amount of information transmitted from attention heads to MLPs is greater in larger models. We now express the core ingredients of this theory in a simplified mathematical model. 
\\

\noindent \textit{A minimal model of meta-memorization.} In this model, a set of $H$ heads learn to convey information to a dense associative memory (DAM) for next-token prediction~\citep{ramsauer2020hopfield}. We consider a dataset with $K$ Bernoulli-binary sequences of length $N$, $\xi^{(1)}, \xi^{(2)}, \ldots, \xi^{(K)}$. $H$ is much smaller than $N$, thus creating an attentional bottleneck. The sequence $\sigma$ is drawn with probability $p_{\sigma}$. We denote $\xi_{n}^{(\sigma)}$ as the state of the $n$-th bit in the $\sigma$-th sequence. 

The DAM has a capacity to store at most $M$ binary sequences of length $N$ (where $M\gg K$). 
During training, the DAM stores the keys to these $M$ sequences $\chi_1,\chi_2,\dots,\chi_M$ as $H$-dimensional vectors $\varphi_1,\varphi_2,\dots,\varphi_M$,
where $\varphi_{\mu} = B \chi_{\mu}$  for $\mu = 1,2,\dots, M$. $B$ is a learnable $H \times N$ matrix. The state of the DAM at a particular time $t$ is summarized by the set of sequences stored in it, $\mathcal{M}_t = (\chi^{(1)},\chi^{(2)},\dots,\chi^{(M)})$.  The DAM is structured as a queue. Given an $N$-bit sequence $\zeta$ during training (which could be a sequence already in $\mathcal{M}_t$), the DAM is updated by adding $\zeta$ as the last element. If the queue has already stored $M$ elements, the first element is removed when $\zeta$ is added. 

During retrieval, the model is given a partial cue $\zeta_{0:n}$ (i.e. the first $n$ bits in the sequence) and has to predict the $n+1$-th bit. To do this, the model computes the $H$-dimensional vector $\hat{\varphi} = \sum_{i=1}^n A_{i}(n) \zeta_i$,
where $A(n)$ is an $H \times n$ matrix such that $A_i(n) \ge 0$ and whose rows sum to 1. 
Thus, one can view $(A_1(n), A_2(n), \dots, A_n(n))$ as an attention map specific to position $n$ and $\hat{\varphi}$ as the query to the DAM. Note that, as in a transformer, the attention maps in general depend on the length of the sequence unmasked thus far. Given the cue $\zeta_{0:n}$ and inverse temperature $\eta>0$, we define the probability of retrieving sequence $\mu$ from the DAM as $$\pi_{\mu} = \frac{e^{\eta \hat{\varphi}^T \varphi_{\mu} }}{\sum_{\mu'} e^{\eta \hat{\varphi}^T \varphi_{\mu'} }} = \frac{e^{\eta \zeta_{0:n}^T A(n)^T B \chi_{\mu} }}{\sum_{\mu'} e^{\eta \zeta_{0:n}^T A(n)^T B \chi_{\mu'}  }}.$$  
The probability that the model assigns to the $n+1$th bit being $+1$ is given by $P(\zeta_{n+1}= +1 | \zeta_{0:n}) = \sum_{\mu \in M_{n+1}^{+}} \pi_{\mu}$
where $M_n^{+}$ is the set of sequences in the DAM that have $+1$ at the $n$-th position.

Intuitively, the model first computes a posterior distribution $\pi$ over sequences in the DAM based on their match with the observed partial sequence $\zeta_{0:n}$, and predicts the next bit by computing the posterior predictive distribution using only sequences that have a $+1$ bit in the $n+1$-th location. Thus, we posit that the MLP layers in the transformer organize during training to implement this two-step inferential procedure. 

Our minimal model has two parameters $A,B$ which are trained by jointly optimizing the loss $\mathcal{L}_t(\sigma)$ at step $t$ for sequence $\sigma$: 
\begin{multline}
\mathcal{L}_t(\sigma; A(n), B, \mathcal{M}_t) = \\
\left( \frac{1}{N} \sum_{n=0}^{N-1} \mathcal {L}_{\text{BCE}}( {\sigma,n} ; A(n), B, \mathcal{M}_t)\right),
\end{multline} where $\mathcal {L}_{\text{BCE}}(\sigma, n; A(n), B, \mathcal{M}_t)$ is the binary cross-entropy loss of predicting the $n+1$th bit of sequence $\xi^{(\sigma)}$ given $A(n), B$ and the current state of the DAM $\mathcal{M}_t$. In our simulations, we compute the loss over a batch of 50 sequences and add these sequences together to the DAM after the update.
\\

\noindent \textit{Meta-memorization in the DAM model.} The DAM model recapitulates the meta-memorization effect. To show this, we performed the data-switch experiment where the training set of sequences is switched halfway to a new set of random sequences, and continued training the matrices $A$ and $B$. The sequences are drawn from a uniform distribution. In Figure \ref{fig:dam}b, we show that accuracy immediately dropped to chance after the switch, but, like Figure~\ref{fig:perf}c, recovered faster than it had at the start of training. The post-switch speedup also increased with scale: models with more heads showed steeper recovery curves. 

Next, we also tested a variant of our DAM where the model uses an InfoNCE objective to learn $A$ and $B$ that maximize the probability $\pi_\mu$ of retrieving the correct sequence $\mu$ given the cue (i.e. partial sequence) $\zeta_{0: n}$, instead of the posterior predictive probability $P\left(\zeta_{n+1}=+1 \mid \zeta_{0: n}\right)$~\citep{oord2018representation}.   We re-ran our experiment where we switched the training set of sequences halfway through training to a fresh set of random sequences and froze the matrices $A$ and $B$ after the switch. The sequences are drawn from a uniform distribution. In Figure \ref{fig:dam}c, we show that even though accuracy dropped after the switch, it remained comparable to the pre-switch accuracy. Thus, the attention heads learn an efficient, dataset-invariant coding scheme for conveying information to the DAM. 
\begin{figure}[t!]
    \centering
    \includegraphics[width=\linewidth]{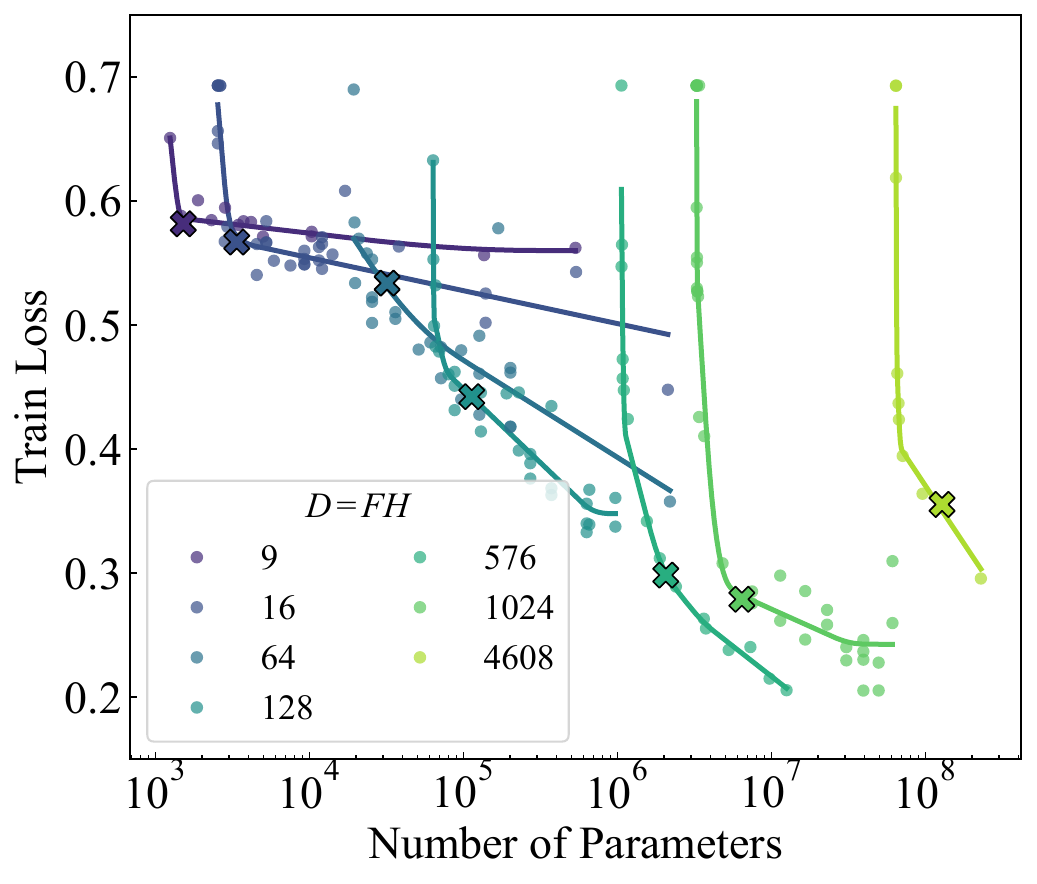}
    \caption{Train loss across a wider parameter sweep across 3 hyperparameters: number of attention heads $H$, width per head $F$, and width $W$ of MLP for $5\times 10^4$ iterations. Each color represents  models with a fixed model dimension $D = FH$ but different $W$. The loss of models with $W = D$ is plotted with cross marks. 
    The elbow of the loss for a fixed $D$ begins approximately at $W \approx D$, suggesting an optimal allocation of parameters to attention heads and MLPs.}
    \label{fig:pareto_front}
\end{figure}
\\

\noindent \textit{The frontier of the scaling law.}  
To understand the extent to which architectural details matter beyond a simple parameter count, we trained 1-layer models with a grid search over number of attention heads $H$, width per head $F$, and MLP width $W$, for $5 \times 10^4$ iterations (Figure~\ref{fig:pareto_front}), allowing us to trade off attention and the MLP parameters  for a given parameter budget. For a fixed model dimension $D = FH$, we observe that performance degrades substantially when $W < D$, suggesting the MLP is the bottleneck. When $W > D$, additional width $W$ yields only marginal improvements, suggesting the MLP is no longer the limiting factor. The elbow at $W \approx D$ (for a fixed $D$) indicates how parameters should be allocated to attention heads and MLPs to minimize loss given a fixed parameter budget. 

\section{Conclusion}
By systematically quantifying memorization in models of different architectures and scales, we find that memorization capacity scales primarily with parameter count, recapitulating key scaling laws observed in models trained on language and synthetic data~\citep{tirumala2022memorization, carlini2022quantifying, morris2025much,li2026incompressible}, and  obtain a scaling law of memorization rate with parameter count. The faster memorization of larger models reflects a meta-memorization effect, where the model's memorization of sequences \textit{in general} accelerates after memorizing other, unrelated, sequences. While the memorization rate \(r(t)\) follows a common scaling law with training iterations \(t\), the upper cutoff of the scaling law increases with model size. 

Mechanistic experiments in one-layer transformers suggest a division of labor between attention and feedforward blocks. MLPs store sequence-specific associations, whereas attention heads learn reusable, position-invariant representations of partial contexts that make new sequences easier to write and retrieve. We develop a minimal model of this division of labor using a dense associative memory with an attentional bottleneck. Together, these results suggest that transformers acquire general-purpose attentional mechanisms that make each incremental gradient update more effective at writing memories and offer a mechanistic account of why scale accelerates memorization.

Our definition of memorization relies on a fixed per-sequence accuracy threshold and on rank-based summaries of the memorized set. Although these choices reveal robust trends across model scale, the precise capacity and rate estimates may depend on the threshold, sequence length, optimizer, learning rate, sampling distribution, and architecture.

One limitation of our work is our focus on memorization of random binary sequences. LLMs in realistic settings need to balance memorization of arbitrary information with acquisition of reusable linguistic structure. A natural direction for future work is to extend the present setting to data distributions with controlled amounts of structure, such as Markov chains or latent-variable sequence families. Such experiments could determine whether meta-memorization is strengthened or weakened when memorization competes with reusable abstraction. Finally, a complete theory should derive the rate exponent and its scale-dependent cutoff from architectural features such as the number of heads, feature dimension, memory capacity, and context length. 

\section*{Acknowledgments}
We thank members of the Reddy lab for insightful comments. GR was partially supported by a joint research agreement between NTT Research Inc. and Princeton University. The simulations presented in this article were performed on computational resources managed and supported by Princeton Research Computing, a consortium of groups including the Princeton Institute for Computational Science and Engineering (PICSciE) and Research Computing at Princeton University. 

\bibliography{references}

\clearpage
\appendix
\onecolumngrid
\renewcommand{\thefigure}{S\arabic{figure}}
\renewcommand{\theHfigure}{S\arabic{figure}}
\setcounter{figure}{0} 

\section{Additional figures}

\begin{figure}[!htbp]
     \centering
     \includegraphics[width=0.95\textwidth]{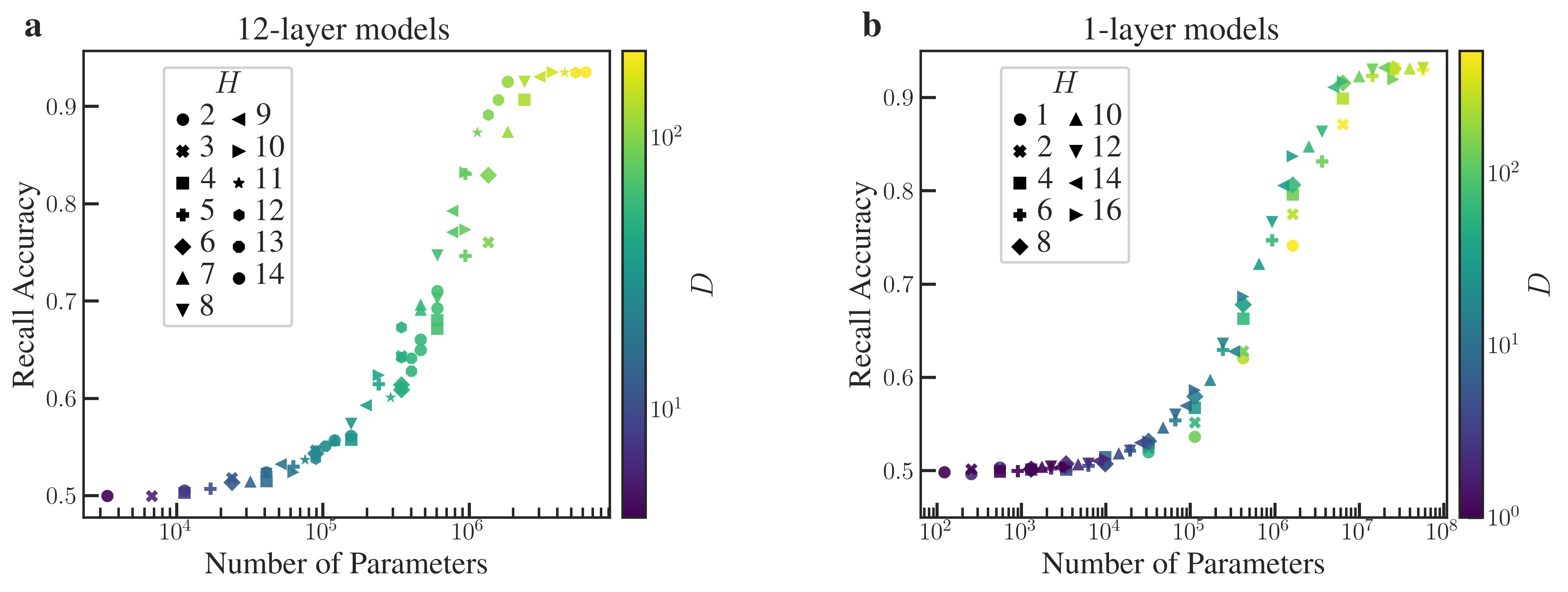}
     \caption{Recall accuracy scales with parameter count. Average next-bit accuracy on the top $10000$ sequences increases with total parameters across architectures that vary in $H$ and $D$.}
     \label{fig:acc_scaling} 
 \end{figure}

\begin{figure}[!htbp]
     \centering
     \includegraphics[width=0.95\textwidth]{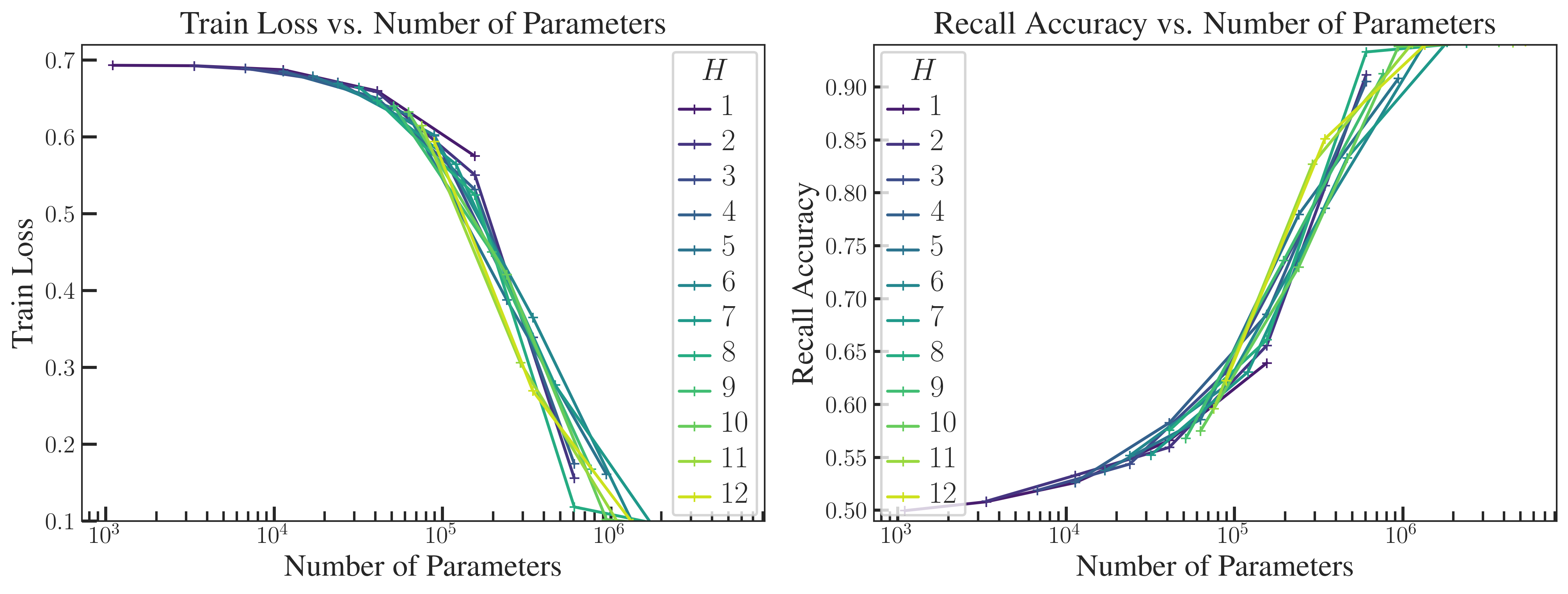} 
     \caption{Train loss and recall accuracy versus total parameter count for 12-layer GPTs trained on binary sequences, sampled uniformly with replacement. Models are trained on $K=10{,}000$ sequences with context length $100$ over $10^5$ training iterations. We vary the number of attention heads $H$ and the number of hidden features per head $F$. Train loss decreases and recall accuracy increases with total parameter count, mostly independently of architecture.}
     \label{fig:uniform_loss_accuracy} 
 \end{figure}

\begin{figure}[!htbp]
     \centering
     \includegraphics[width=0.65\textwidth]{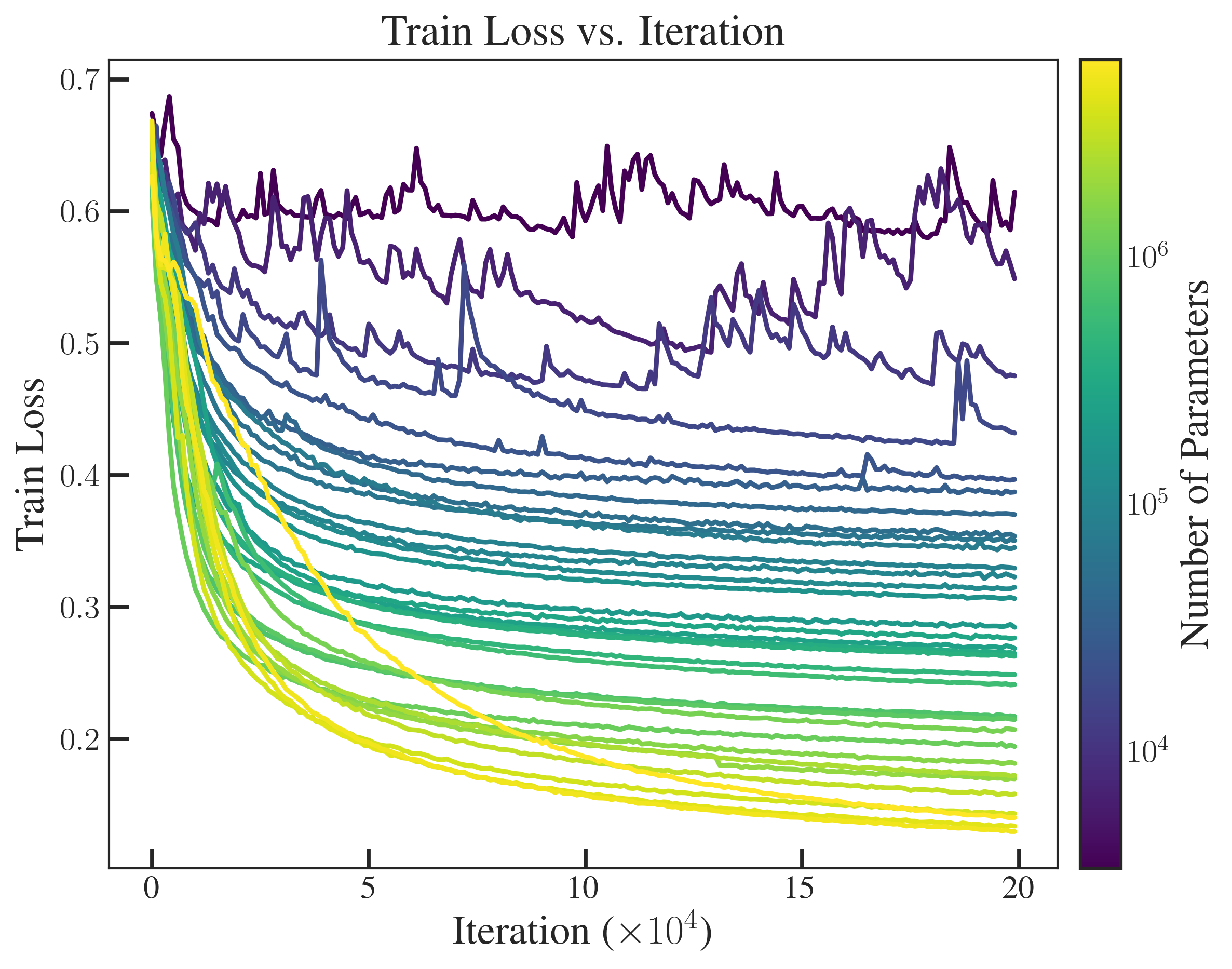}
     \caption{Training loss dynamics across model scales for 12-layer models trained on Zipfian distributed sequences. Larger models generally reach lower loss, even though the largest models show increased loss compared to mid-size models since we did not explicitly tune our optimization hyperparameters.}
     \label{fig:loss_dyn} 
 \end{figure}

\begin{figure}[!htbp]
     \centering
     \includegraphics[width=0.95\textwidth]{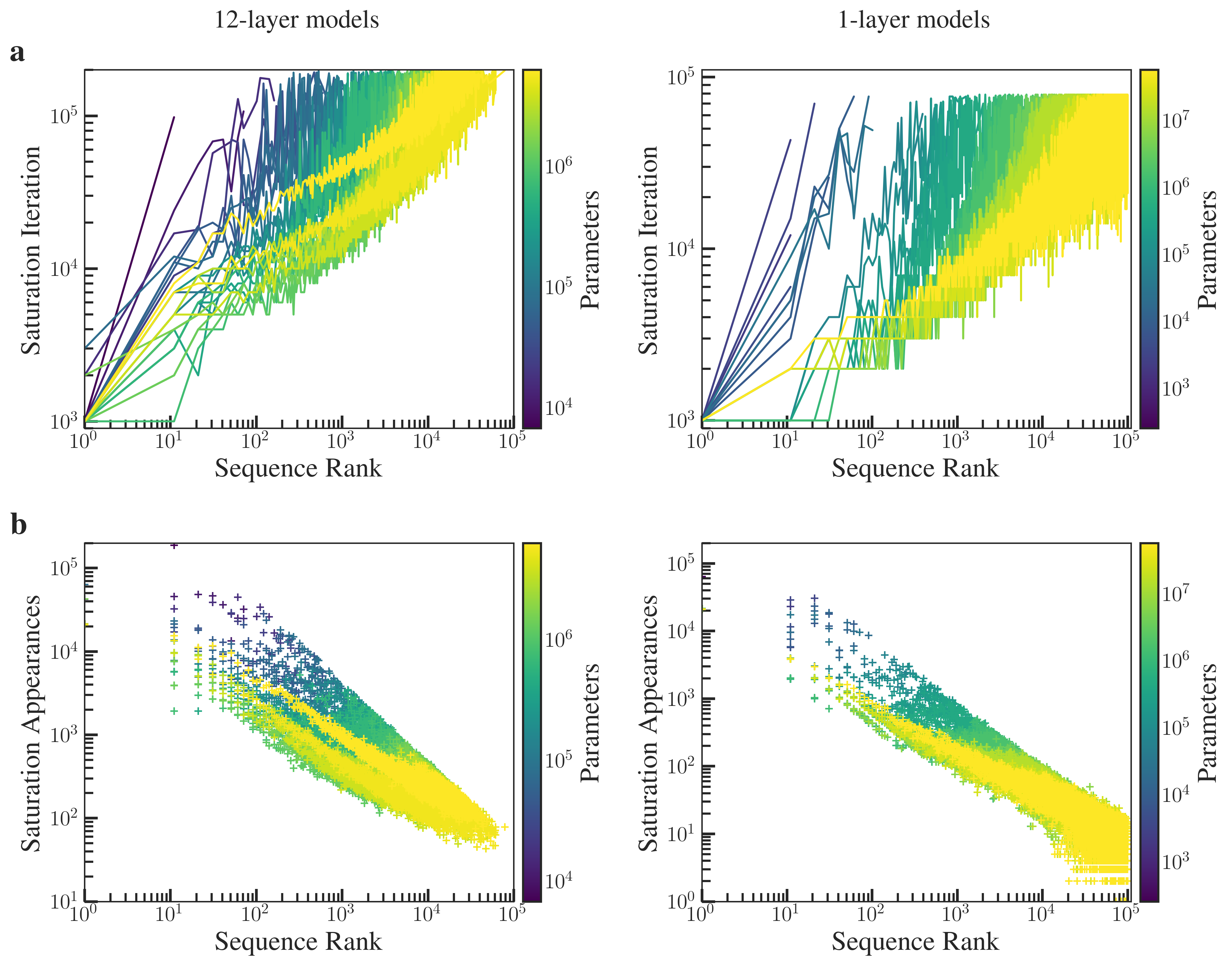}
     \caption{Meta-memorization in 12-layer and 1-layer models. (a) Saturation iteration vs. sequence rank for 12-layer and 1-layer models. We define a sequence's saturation iteration as the iteration the model ``memorized'' it (i.e. attains average accuracy above 90\%). We observe that sequences of higher rank are learned later in time, mainly because they are presented less frequently due to the Zipfian data distribution. However, by taking a vertical slice through our figures, we see that larger models memorize faster, since for any given sequence rank, larger models generally require fewer iterations to memorize it. By taking a horizontal slice through our figures, we see that at any given time, larger models will generally have learned more sequences. (b) Appearances at saturation vs. sequence rank for different model sizes. We define appearances at saturation as the number of presentations $n_k$ the model takes to ``memorize'' a sequence  $k$ (i.e. learned to 90\% accuracy). We observe that higher-ranked sequences require fewer presentations to be memorized. Interestingly, we observe that the very largest 1-layer models require as few as 2 presentations to memorize a sequence (bottom right).}
     \label{fig:saturation_raw} 
 \end{figure}

\begin{figure}[!htbp]
     \centering
     \includegraphics[width=1\textwidth]{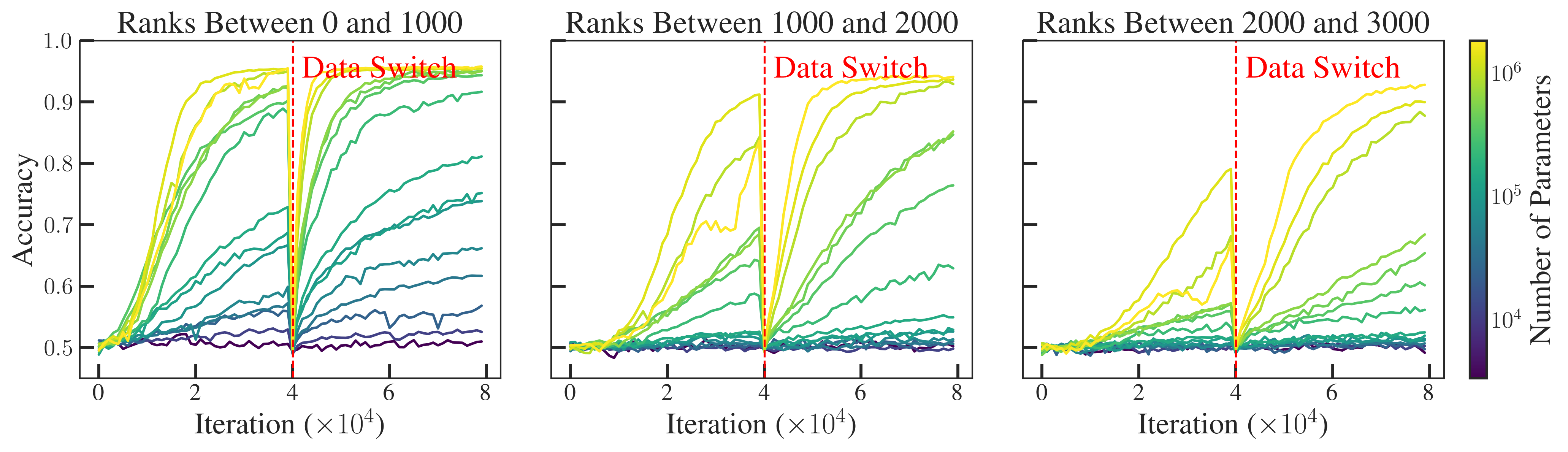}
     \caption{Meta-memorization in data-switching experiments for 12-layer models across sequences of different ranks. We switched the training set of sequences after $4\times 10^4$ iterations to a fresh set of random sequences drawn from the same Zipfian distribution. For sequences across different ranks, accuracy immediately returned to chance after the switch, but recovered faster relative to the start of training (See Figure~\ref{fig:perf}c).}
     \label{fig:12layer_data_switch_3_panels} 
 \end{figure}

\begin{figure}[!htbp]
     \centering
     \includegraphics[width=1\textwidth]{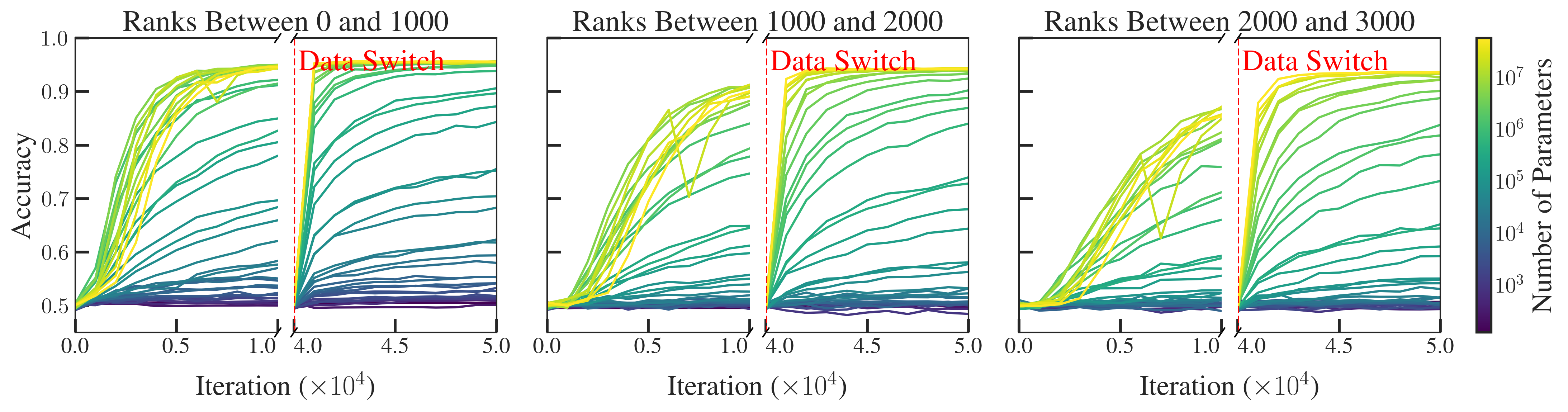}
     \caption{Meta-memorization in data-switching experiments for 1-layer models across sequences of different ranks. We switched the training set of sequences after $4\times 10^4$ iterations to a fresh set of random sequences drawn from the same Zipfian distribution. For sequences across different ranks, accuracy immediately returned to chance after the switch, but recovered faster relative to the start of training (See Figure~\ref{fig:perf}c). Due to the very rapid learning in 1-layer models, here we show results only for iterations $[0,10^4]$ and $[4\times10^4,5\times 10^4]$ to better illustrate the effect.}
     \label{fig:OL_data_switch_3_panels} 
 \end{figure}

\begin{figure}[!htbp]
     \centering
     \includegraphics[width=.5\textwidth]{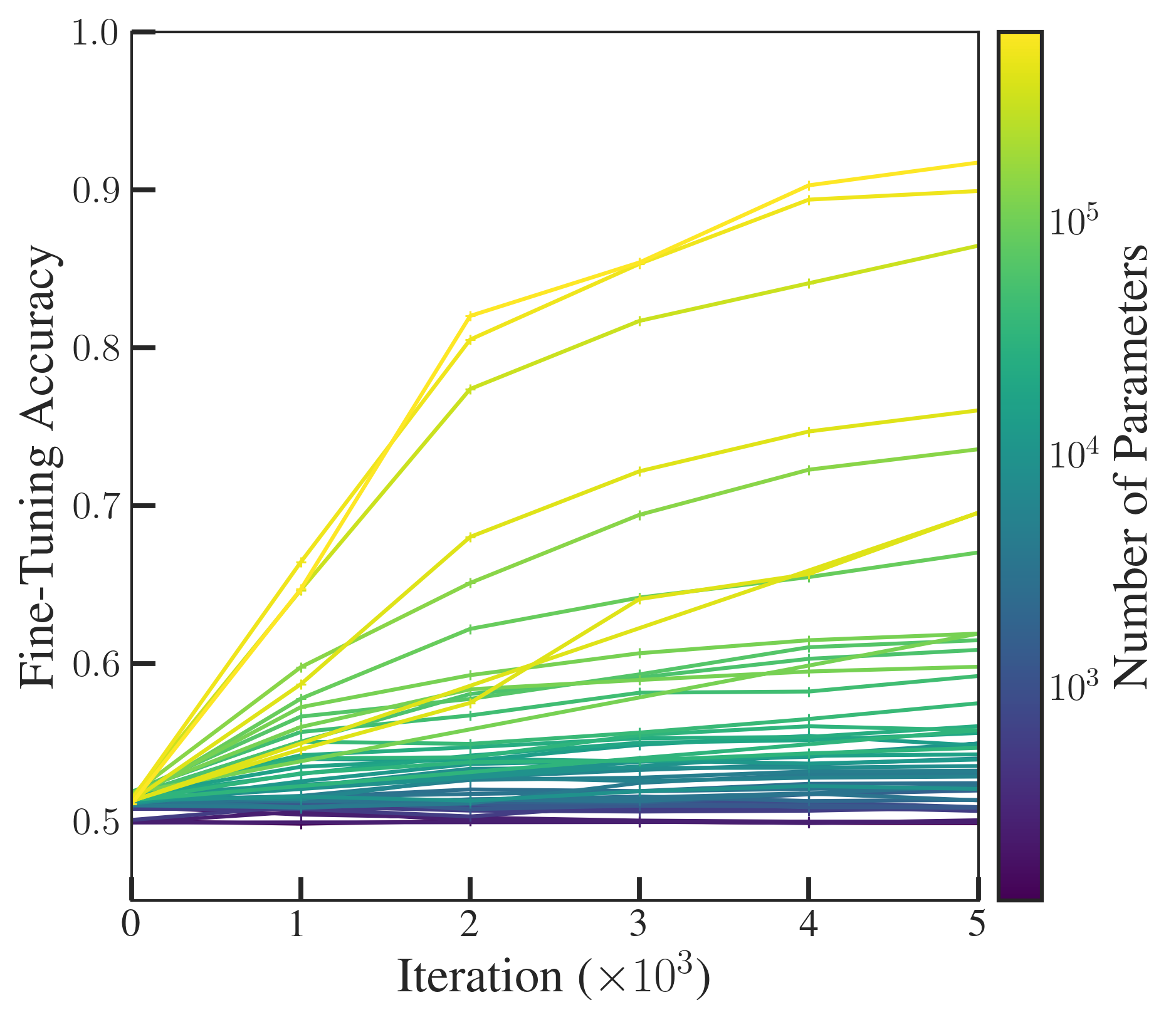}
     \caption{Fine-tuning accuracy of 1-layer transformers on novel unseen sequences increases as a function of training. Larger models achieve higher fine-tuning accuracy.}
     \label{fig:finetuning_accuracy_vs_time} 
 \end{figure}

\begin{figure}[!htbp]
     \centering
     \includegraphics[width=.5\textwidth]{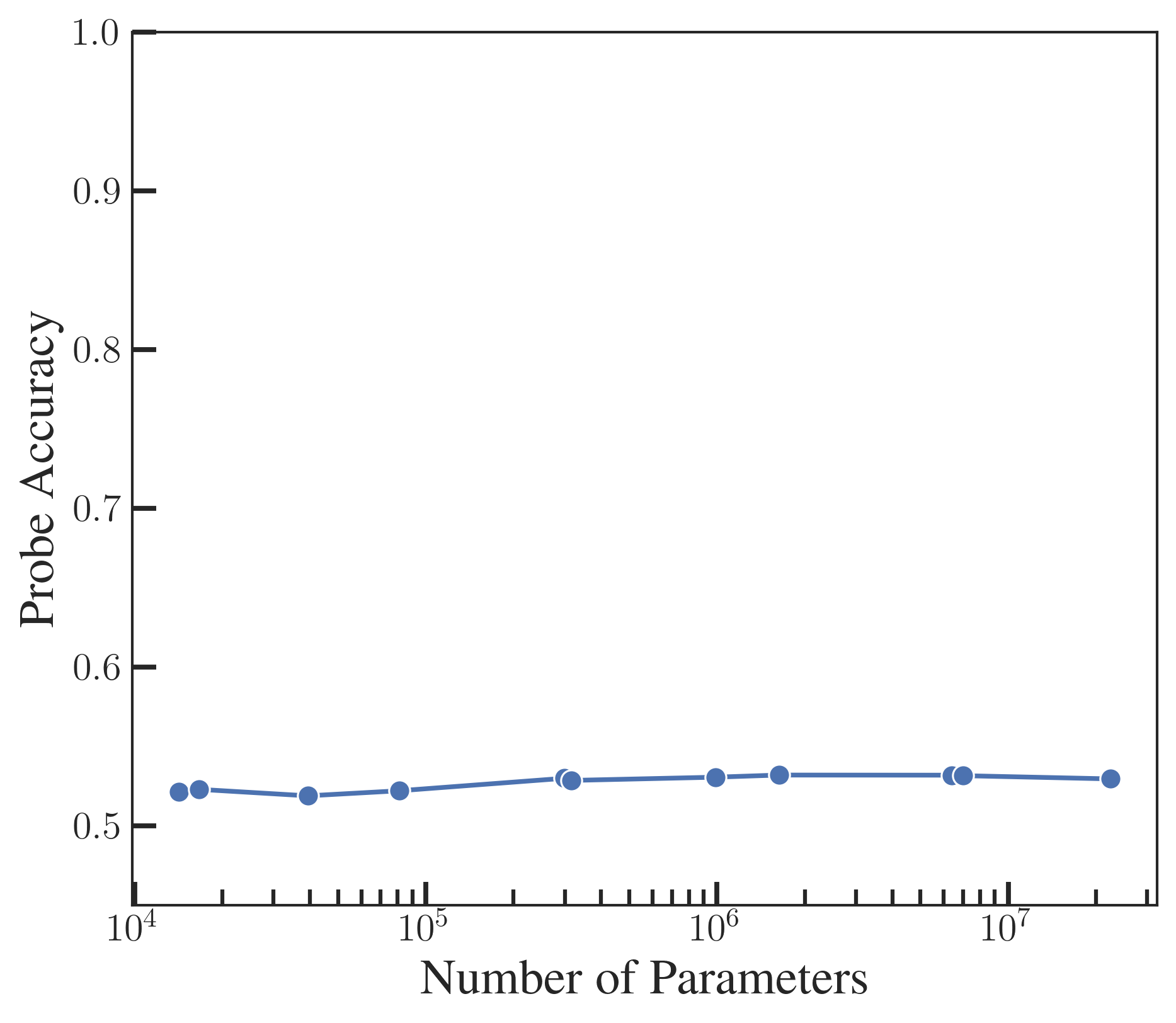}
     \caption{1-layer models with attention weights set to zero throughout training and probing fail to learn novel sequences.}
     \label{fig:zero_attn_weights} 
 \end{figure}

\begin{figure}[!htbp]
     \centering   \includegraphics[width=0.95\textwidth]{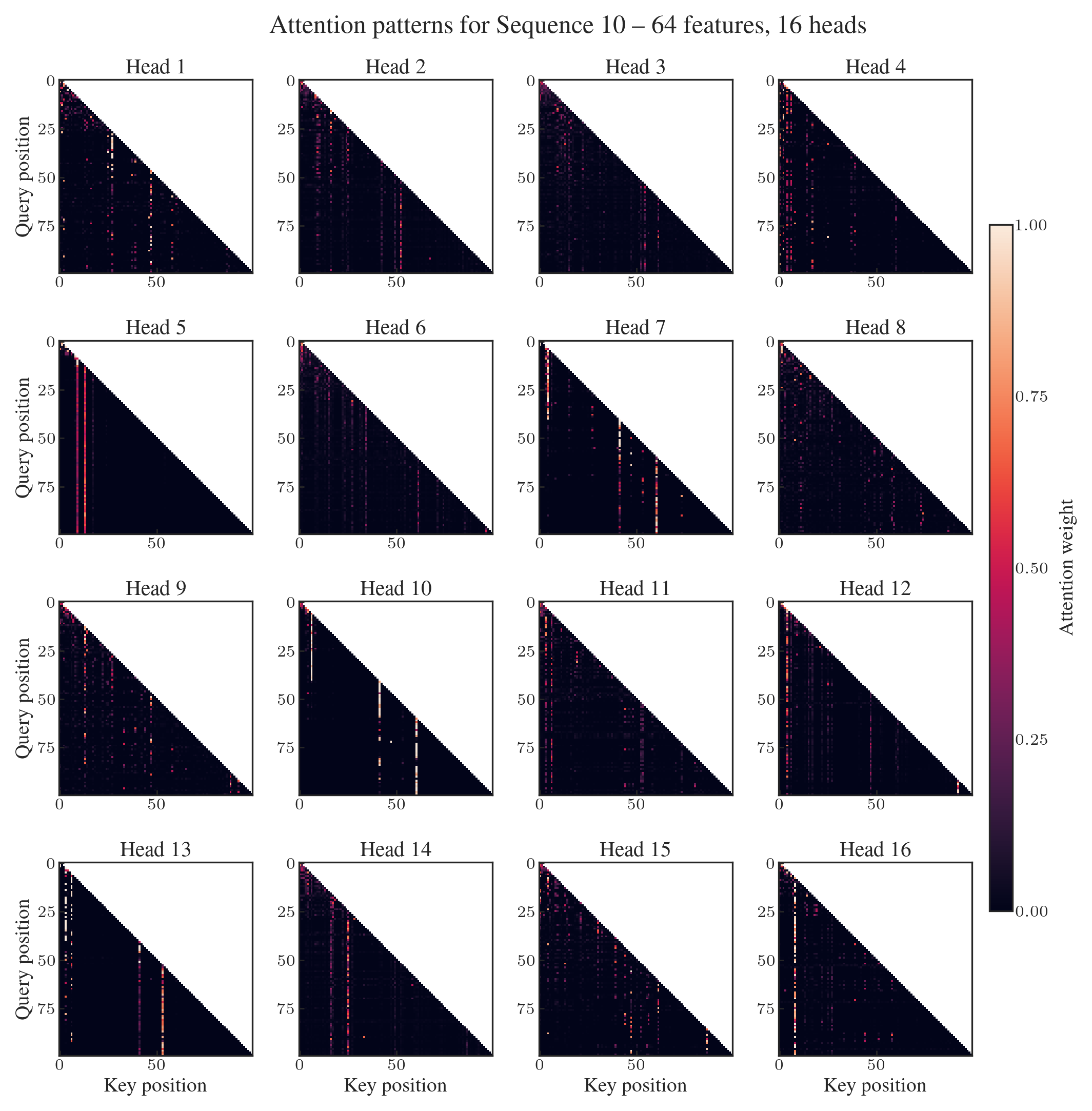}
     \caption{Learned attention maps in a trained one-layer transformer for Sequence 10 (16 heads, 64 features per head). Each panel shows attention maps of one head, evaluated on the that sequence. Rows are query positions and columns are causal key positions.}
     \label{fig:attention_weights_sequence_10} 
 \end{figure}

\begin{figure}[!htbp]
     \centering
     \includegraphics[width=.95\textwidth]{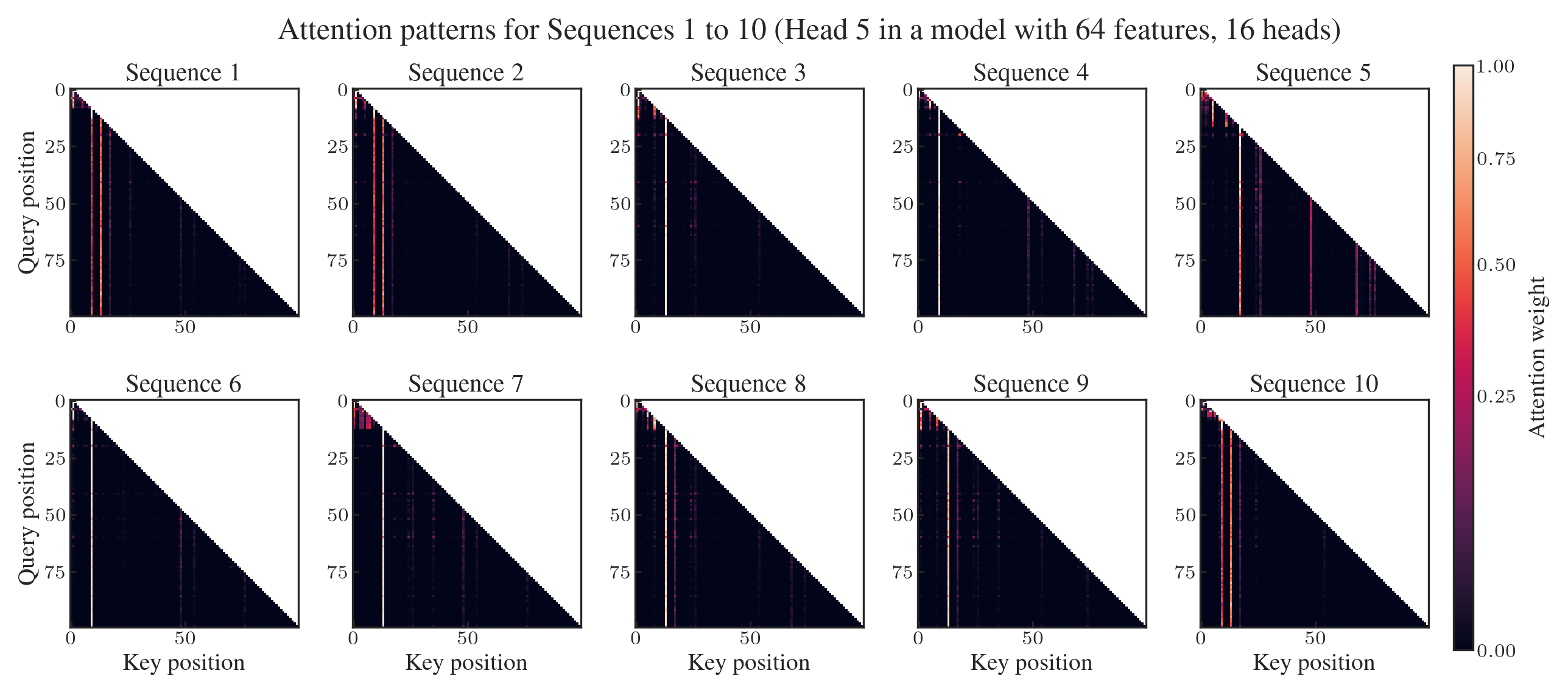}
     \caption{Attention maps for the fifth head in a trained one-layer transformer across multiple sequences (16 heads, 64 features per head). Each panel shows the attention maps for this head, evaluated on the 10 lowest-ranked sequences. Rows are query positions and columns are causal key positions. We observe vertical bands at similar positions across multiple  random sequences.}
     \label{fig:attention_weights_head_5} 
 \end{figure}

\end{document}